\documentclass[conference]{IEEEtran}

\usepackage{amsmath,amssymb}
\usepackage{graphicx}
\usepackage{hyperref}
\usepackage{cite}
\usepackage{float}
\usepackage{multirow}
\usepackage{xcolor}
\usepackage{booktabs}

\title{Universal Differential Equations for Scientific Machine Learning of Node-Wise Battery Dynamics in Smart Grids}

\author{%
  \IEEEauthorblockN{Tarushri N.~S.}%
  \IEEEauthorblockA{Indian Institute of Science, Bengaluru, India\\
    Email: tarushrin@iisc.ac.in}%
}

\begin{document}

\maketitle

% ------------------ Abstract ------------------
\begin{abstract}
Universal Differential Equations (UDEs), which blend neural networks with physical differential equations, have emerged as a powerful framework for scientific machine learning (SciML), enabling data-efficient, interpretable, and physically consistent modeling. In the context of smart grid systems, modeling node-wise battery dynamics remains a challenge due to the stochasticity of solar input and variability in household load profiles. Traditional approaches often struggle with generalization and fail to capture unmodeled residual dynamics.

This work proposes a UDE-based approach to learn node-specific battery evolution by embedding a neural residual into a physically inspired battery ODE. Synthetic yet realistic solar generation and load demand data complete with diurnal structure, noise, and node-wise variation are used to simulate battery dynamics over time. The neural component learns to model the unobserved or stochastic corrections that arise from heterogeneity in node demand and environmental conditions. Key contributions include the construction of node-aware load profiles, the use of low-frequency noise to ensure temporal diversity, and the demonstration of temporal extrapolation capability.

Comprehensive experiments reveal that the trained UDE successfully aligns with ground truth battery trajectories, exhibits smooth convergence behavior, and maintains stability in long-term forecasts. These findings affirm the viability of UDE-based SciML approaches for battery modeling in decentralized energy networks, and suggest broader implications for real-time control and optimization in renewable-integrated smart grids.

\textit{Index Terms}\textemdash Deep learning, neural networks, universal differential equations, smart grids, machine learning, battery dynamics, energy forecasting.
\end{abstract}

\section{Introduction}
Scientific Machine Learning (SciML) aims to combine the strengths of mechanistic physical modeling and data-driven learning. Within this paradigm, Universal Differential Equations (UDEs) have emerged as a hybrid modeling approach that incorporates known physical laws while learning the unknown or unmodeled components from data\cite{rackauckas2020ude}. UDEs are especially promising in domains where partial physical knowledge exists, and data can inform corrections or refinements.

Recent studies highlight the potential of physics-informed neural networks (PINNs) in power grid modeling and control \cite{PINN_review}. Smart grids, with their growing reliance on distributed renewable energy sources and local battery storage, present an ideal application area for UDEs. Accurately modeling battery charge and discharge dynamics is crucial for grid reliability and optimization, yet conventional approaches often fall short due to their inability to handle local load variability, uncertainty in solar input, and missing model terms. While deep learning models offer flexibility, they often require vast amounts of training data and suffer from poor interpretability and extrapolation.

This study addresses the gap between rigid physical models and black-box neural networks by applying UDEs to the problem of battery modeling in a multi-node smart grid. Each node in the system is equipped with a battery that responds to solar input and load demand. We use a base physical equation for battery evolution and introduce a neural network to model residual dynamics caused by heterogeneity in consumption and unmeasured influences.

The key questions guiding this study are:
(i) Can UDEs accurately learn residual battery dynamics from synthetic solar and load data with noise? (ii) How well does the learned model generalize to extended forecasting periods? (iii) Can small node-level variations be captured through shared physical structure with individualized learned corrections?

The goal is to demonstrate the potential of UDEs to serve as interpretable, generalizable, and scalable models for decentralized energy systems, with applications in forecasting, control, and planning.

\section{Methodology}
\subsection{Synthetic Data Generation}

To simulate realistic battery behavior, we generate synthetic solar and load demand signals that evolve over time, mimicking real-world variability. Both are defined as continuous functions of time and sampled hourly over a 10-day window.

The solar power profile is modeled as a diurnal half-sine function with additive low-frequency noise:
\[
P_s(t) = \max\Bigl(0, \sin\bigl(\tfrac{\pi\,(t \bmod 24) - 6}{12}\bigr)\Bigr) + \epsilon_s(t),
\]
where
\[
\epsilon_s(t) = 0.05\,\sin(0.1\,t)
\]
introduces smooth weather-like fluctuations.

Each node’s load demand signal combines a baseline and Gaussian peaks for morning and evening usage:
\[
P_d^{(i)}(t) = b_i + 0.3\,\exp\Bigl(-\tfrac{(t \bmod 24 - 8)^2}{2\,\sigma^2}\Bigr)
\]
\[
 + 0.4\,\exp\Bigl(-\tfrac{(t \bmod 24 - 19)^2}{2\,\sigma^2}\Bigr) + \epsilon_d^{(i)}(t),
\]

with node-specific base values $b_i \in \{0.45,0.5,0.55\}$, peak width $\sigma=1.5$, and
\[
\epsilon_d^{(i)}(t) = 0.02\,\sin(0.07\,t + \phi_i),
\]
adding load variability across nodes.

The term $(t \bmod 24)$ ensures this cycle repeats every 24 hours, mimicking the natural progression of sunrise to sunset. The sine wave peaks at midday (around hour 12) and drops to zero at night. To make the simulation more realistic, we add a low-frequency sinusoidal noise component. This represents smooth weather-driven variability, such as transient cloud cover or atmospheric haze. The noise ensures that each day's solar curve is slightly different from the others, even though the overall pattern remains consistent.

The two Gaussian peaks occurs once in the morning (centered around 8 AM) and once in the evening (around 7 PM). These reflect typical residential usage patterns, for example, people preparing for the day or returning home after work. The width of these peaks is controlled to simulate concentrated versus spread-out consumption. Again, smooth low-frequency noise is added to each node's demand signal to create small variations over time. Furthermore, each node has a slightly different base demand level to reflect consumer heterogeneity.
Together, these synthetic inputs generate complex, time-varying net power flows at each node. The solar-load mismatch drives the battery to charge when excess solar power is available and discharge when demand exceeds supply. This interplay produced rich dynamics that serve as inputs to the Universal Differential Equation model during training.

\begin{figure}[H]
  \centering
  \includegraphics[width=0.48\textwidth]{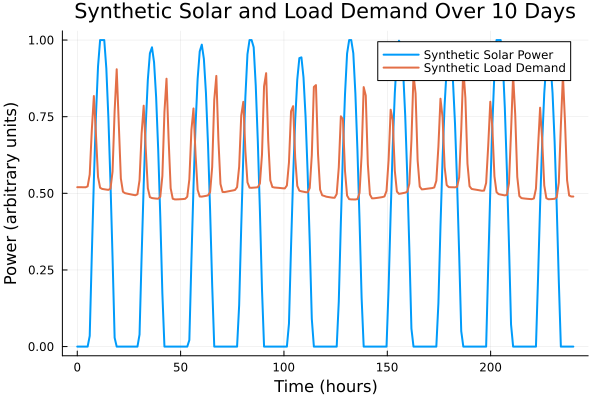}
  \caption{Synthetic solar and load demand over the 10-day period. 
  The plot exhibits the synthetic solar and load demand inputs used to simulate battery dynamics. Solar power follows a diurnal half-sine curve modulated by low-frequency noise, simulating real-world variability. The load demand combines a baseline level with Gaussian peaks. This diverse structure drives complex battery behavior across nodes.}
  \label{fig:demand10d}
\end{figure}

Figure~\ref{fig:demand10d} shows the resulting solar and load profiles over the full 10-day training window.

\noindent

\begin{figure}[H]
  \centering
  \includegraphics[width=0.48\textwidth]{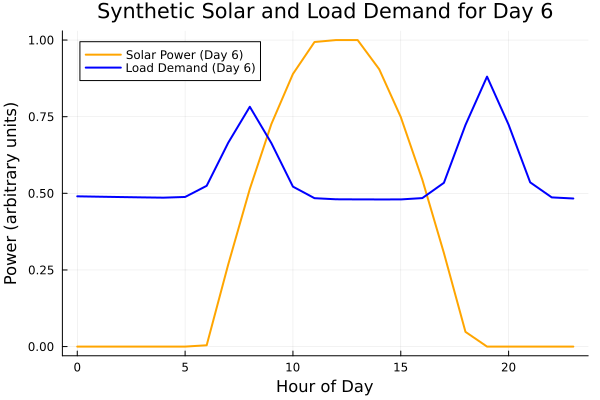}
  \caption{Solar power and load demand for Day~6.
  Here, solar power peaks around midday while two demand spikes occur in the morning and evening, driving predictable charge–discharge cycles.}
  \label{fig:day6}
\end{figure}

Figure~\ref{fig:day6} provides a detailed snapshot of solar and load patterns on Day 6. Demand shows two sharp peaks, while solar production is concentrated around noon. These temporal mismatches induce predictable charging during sunlight hours and discharge around consumption peaks, behavior the model must learn to replicate.

\noindent

\subsection{Physical Battery Model}
The base model for battery evolution is formulated using an energy balance principle. For each node $i$, we define the battery state $E_b^{(i)}(t)$ as the cumulative stored energy in the battery at time t, measured in arbitrary units. The rate of change of this energy is governed by the difference between power input from solar and the power consumed by the node's load.
The base battery energy evolution for node $i$ follows:
\[
\frac{dE_b^{(i)}}{dt} = P_s(t) - P_d^{(i)}(t).
\]
where $P_s(t)$ is the solar power input (common to all nodes) and $P_d^{(i)}(t)$ is the load demand specific to node $i$.
This differential equation assumes ideal charge/discharge efficiency, no storage losses, and no upper or lower bounds on storage capacity.
The interpretation is straight forward:
(i) When $P_s(t) > P_d^{(i)}(t)$, the battery charges, increasing $E_b^{(i)}(t)$. 
(ii) When $P_s(t) < P_d^{(i)}(t)$, the battery discharges, decreasing $E_b^{(i)}(t)$. 
(iii) The instantaneous net power governs the slope of $E_b^{(i)}(t)$, and its integral over time defines the battery state.

\begin{figure}[H]
  \centering
  \includegraphics[width=0.48\textwidth]{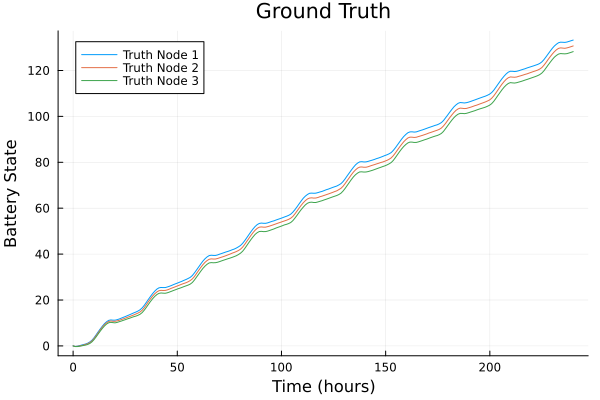}
  \caption{Ground truth battery state evolution for all nodes over 10 days.
  Each curve corresponds to a different node with a distinct baseline demand. The battery state generally increases during daytime hours when solar input exceeds load, resulting in upward slopes. Conversely, during night or high-demand hours, the battery discharges, leading to plateaus or slight declines. The trajectory exhibits staircase-like pattern, reflecting alternating periods of charge and discharge over each 24-hour cycle.}
  \label{fig:truth}
\end{figure}

To validate the synthetic solar and demand inputs, we simulate the battery evolution across all three nodes using this physical model without any learned neural residual.
Figure~\ref{fig:truth} illustrates the staircase-like accumulation over time.
The differences among the three nodes are due to their base load shifts: nodes with higher demand accumulate energy more slowly, resulting in lower overall battery states. This setup creates physically plausible, heterogeneous trajectories for training and evaluating the UDE model.

\subsection{Universal Differential Equation Formulation}
While the base ODE captures the net physical behavior of the battery state driven by solar input and load demand, it may fail to account for system imperfections, latent dynamics, or environmental factors that are not explicitly modeled. To address this, we extend the battery evolution equation using the Universal Differential Equation (UDE) framework, which integrates a learnable neural network residual into the dynamics.
To model unobserved dynamics and correct for residuals, we augment the base ODE with a neural network term:
\[
\frac{dE_b^{(i)}}{dt} = P_s(t) - P_d^{(i)}(t) + \mathrm{NN}_\theta(t, E_b^{(i)}),
\]
where $\mathrm{NN}_\theta$ is a fully connected feedforward network parameterized by weights $\theta$, learning node-specific corrections from time and state inputs, and its output represents a data-driven correction to the physical model, capturing residual patterns not explained by solar and load terms.
This neural residual allows the model to learn unobserved influences such as behavioral variability in consumption not reflected in the base demand profiles.

By combining a known physical structure with a flexible learning component, the UDE preserves interpretability while improving expressiveness. During training, the weights $\theta$ are optimized to minimize the error between predicted and true battery trajectories across all nodes, enabling the network to model node-specific corrections that adapt over time and system state.

\noindent

\subsection{Network Architecture}
The neural network component $NN_\theta(t,E_b)$ is implemented as a feedforward multilayer perceptron. It accepts a two-dimensional input vector composed of the current simulation time $t$ and battery state $E_b$, both normalized. The input is first passed through a hidden layer of 16 neurons with hyperbolic tangent (tanh) activation, followed by a second hidden layer of 16 neurons using the same nonlinearity. The final output layer is a single neuron with a linear activation that produces the residual correction term.
This architecture is intentionally lightweight to balance expressiveness with computational efficiency. The use of tanh activations allows the network to model smooth and bounded corrections, consistent with physical expectations. 

\subsection{Training and Optimization}
We train the Universal Differential Equation (UDE) model using the Julia-based SciML ecosystem, primarily relying on the \texttt{Lux.jl}, \texttt{DiffEqFlux.jl}\cite{diffEqFluxjl}\cite{rackauckas2017diffeq} and \texttt{Optimization.jl}\cite{optimizationjl} packages. The neural network that defines the residual term is constructed and managed using \texttt{Lux.jl}, a high-performance, lightweight neural network library designed for scientific computing\cite{luxjl}. The goal is to optimize the parameters $\theta$ of the neural network embedded in the ODE so that the predicted battery dynamics closely match the ground truth trajectories. The training uses automatic differentiation (AD) to compute gradients through the solution of the ODE\cite{rackauckas2021adjoint}. Specifically, we use the continuous adjoint sensitivity method for efficient backpropagation through the ODE solver. It is efficient and memory-scalable for long time horizons. AD is handled via \texttt{Zygote.jl}\cite{zygotejl} in conjunction with the \texttt{InterpolatingAdjoint} method provided by SciMLSensitivity.\cite{rackauckas2021adjoint}

 Network weights are initialized randomly, and training is performed using the \textbf{ADAM Optimizer} with a learning rate of $0.005$. The model is trained for 300 iterations, balancing convergence speed with computational efficiency. The loss is the mean squared error across all nodes and time points:
\[
L(\theta) = \sum_{i,t} (E_{b,\mathrm{pred}}^{(i)}(t;\theta) - E_{b,\mathrm{true}}^{(i)}(t))^2.
\]
where: $E_{b,\mathrm{pred}}^{(i)}(t;\theta)$ is the predicted battery state from the UDE model for node $i$ at time $t$, and $E_{b,\mathrm{true}}^{(i)}(t)$ is the corresponding ground truth value from the physical model.
Training proceeds by solving the forward UDE with current weights, computing the error, and updating the weights using ADAM based on the gradient of $L(\theta)$. The model's performance and loss evolution are tracked and visualized, as shown in the Results section.

\section{Results}

\subsection{Training Convergence}
\begin{figure}[H]
  \centering
  \includegraphics[width=0.48\textwidth]{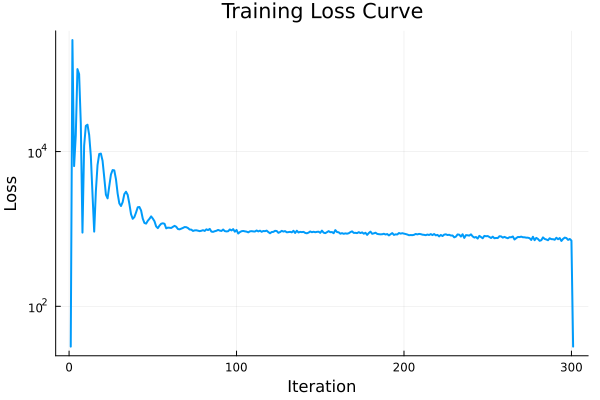}
  \caption{Training loss (log scale) over 300 ADAM iterations.}
  \label{fig:loss}
\end{figure}

The model was trained using the ADAM optimizer for 300 iterations. As shown in the training loss curve (Figure~\ref{fig:loss}), the evolution of the loss, reveals three distinct phases of convergence, each reflecting meaningful transitions in model behavior.
In the initial phase, the loss exhibits a sharp increase. This counterintuitive rise is a well-documented phenomenon in UDE training, often caused by the random initialization of neural network weights. Early in training, these weights may produce residual corrections that worsen the model's mismatch with solar input and load demand, leading to a temporarily destabilized trajectory and higher error. Early in training, as gradient magnitudes are still large, ADAM's adaptive moment estimates can momentarily destabilize the loss before settling into a proper descent\cite{rackauckas2017diffeq}.
The second phase marks the onset of optimizer-driven adaptation. The small oscillations observed in the loss curve arise from the stochastic nature of the ADAM optimizer and its moment-based updates\cite{Kingma2015Adam}. As gradient updates begin to reshape the residual network, the model exits the spurious initial basin and begins aligning its corrections with the actual underlying mismatch. This results in a consistent downward slope in the loss curve as the neural component starts capturing useful residual dynamics.
In the final phase, the training loss plateaus and stabilizes at a low value. This flat region suggests that the neural network has converged to a corrective form that complements the physical model without overfitting. The residual term now robustly accounts for non-modeled variability across time and nodes, such as smooth fluctuations in load and demand, while preserving the dominant structure imposed by the solar-load balance.
Notably, a sharp dip in the final few iterations suggests that the optimizer may have reached a smoother minimum in the loss landscape after a prolonged plateau. This behavior reflects the gradual refinement of residual corrections and the model's ability to converge toward a more optimal parameter configuration even late in training.

Overall, the training trajectory confirms the efficacy of combining physics-based system equations with data-driven adaptation. The sharp decline and stabilization in loss demonstrate that the UDE learned meaningful corrections, enhancing generalization without compromising physical interpretability.

\subsection{Learned Battery Dynamics vs Ground Truth}
\begin{figure}[H]
  \centering
  \includegraphics[width=0.48\textwidth]{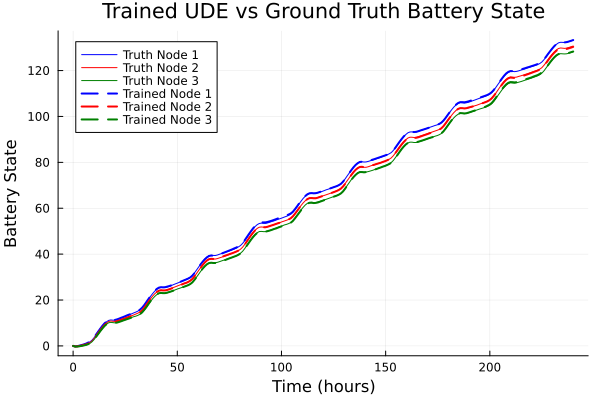}
  \caption{Comparison of UDE-predicted and true battery states across all nodes.}
  \label{fig:trained_vs_truth}
\end{figure}

Figure~\ref{fig:trained_vs_truth} presents a comparison of UDE-predicted battery states against ground truth trajectories generated from purely physical equations. All three nodes demonstrate strong agreement between trained predictions and true dynamics, confirming the UDE's ability to recover underlying structure while compensating for non-modeled effects.
The trained predictions (dashed lines) align closely with the ground truth (solid lines), reproducing the staircase-like evolution that arises from alternating solar charging and load-driven discharge cycles. Importantly, while each node experiences different baseline demand levels, the UDE adapts accordingly and learns node-specific correction terms without overfitting or instability.
This result highlights the UDE's capacity to generalize across heterogeneous dynamics using a shared residual neural structure. The accuracy across all nodes suggests that the network has learned to model consistent non-physical contributions, such as smooth fluctuations or consumptions irregularities, while respecting the dominant physical trends dictated by solar-load balance.

\subsection{Forecasting Performance}
\begin{figure}[H]
  \centering
  \includegraphics[width=0.48\textwidth]{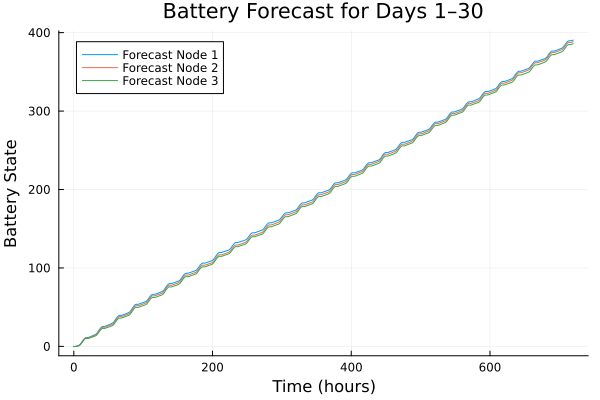}
  \caption{30-day battery forecast for all nodes.}
  \label{fig:forecast}
\end{figure}

To evaluate the generalization ability, the trained UDE was extrapolated from the original 10-day training period to a 30-day forecast horizon. As shown in Figure~\ref{fig:forecast}, the predicted battery states for all three nodes continue to evolve smoothly and realistically throughout the extended simulation window, despite having never seen data from this future range during training.
The forecasted trajectory preserve the key dynamical structure observed in the training phase. The model exhibits no signs of instability, divergence, or error accumulation over time, demonstrating that the learned residual function is not overfit to short-term patterns but instead captures a general correction principle grounded in the physics-informed structure of the system.
The level of stability over a long horizon highlights one of the core advantages of Universal Differential Equations\textemdash  their hybrid design allows neural networks to learn meaningful corrections that are not only locally accurate but also globally consistent. The fact that the predicted dynamics remain bounded, interpretable, and node-specific over 720 hours (30 days) speaks to the robustness of the learned model and its suitability for real-time, long-term control and planning in energy systems.

\subsection{Node-wise Load Sensitivity}
Another fascinating feature of this study is the ability of the UDE to resolve and adapt to node-specific load demand variations, despite being trained on a shared solar input and a single neural architecture. 
Each node was assigned a distinct base load consumption level, slightly higher or lower than the mean, to emaculate heterogeneous energy usage patterns. As a result, even under identical solar conditions, the battery state trajectories diverged subtly across nodes. Nodes with higher baseline demand exhibited slower accumulation of stored energy and lower net battery state over time, while nodes with lower demand experienced faster charging and higher energy retention.
The UDE successfully captured these fine-grained variations through its residual neural term. Rather than fitting a single global pattern, the model learned to modulate its corrections based on the current node state at time, resulting in tailored adjustments that preserve the relative dynamics across nodes. This sensitivity was preserved in both the training and forecasting regimes.
The result highlights the model's capacity to account for distributed, individualized consumption behavior while maintaining structural coherence which is an essential requirement for real-world smart grid applications involving many consumers under shared supply constraints.

\subsection{Summary of Findings}
\begin{itemize}
    \item The UDE model successfully captured battery dynamics by augmenting physics-based system equation with a neural residual term.
    \item It generalized well beyond the 10-day training window, producing stable and realistic forecasts over a 30-day horizon.
    \item The model effectively handled node-specific load variations using a shared neural architecture.
    \item The training process exhibited smooth convergence and no instability.
    \item Overall, the approach demonstrated a robust framework for hybrid modeling in distributed smart grid systems.
\end{itemize}

\section{Discussion}
The successful deployment of a Universal Differential Equation (UDE) framework for modeling node-wise battery dynamics in smart grids highlights the viability of hybrid scientific machine learning approaches in real-world energy applications. Unlike purely data-driven methods, UDEs embed physical structure while retaining flexibility to capture unmodeled influences, a feature crucial for environments with uncertainty in both demand and supply.
Our findings demonstrate that even with modest architectural choices and limited training time, neural augmentations can effectively bridge the gap between modeled and observed behavior. The ability to generalize across time, evidenced by stable 30-day forecasts, supports the idea that such hybrid models can serve as digital twins for smart grid planning and adaptive control.
One of the notable insights was the success of introducing minimal heterogeneity across nodes (via base load shifts) to simulate realistic variation. This enabled the model to learn localized corrections without requiring a redesign of the architecture. The outcome suggests UDEs are scalable to larger grids where buildings, zones, or substations might exhibit varying consumption behavior.
Nevertheless, our approach is not without limitations. The training setup relied on synthetic noise and deterministic solar/load functions. While this was a necessary simplification to demonstrate feasibility, real-world data introduces greater complexity including weather unpredictability, user-driven demand spikes, and battery degradation. Additionally, we only explored first-order battery dynamics; future iterations could consider temperature effects, charge/discharge inefficiencies, and circuit-level behavior.
An important consideration for future research is the integration of control actions. With a learned UDE in place, reinforcement learning or model predictive control schemes could be layered to create autonomous agents capable of real-time decision-making in smart energy environments. Combining physics-informed modeling with data-informed control offers a compelling path forward for resilient and sustainable energy systems.

\section{Conclusion}
In this study, we demonstrated the efficacy of Universal Differential Equations (UDEs) for learning node-wise battery dynamics in a simulated smart grid environment. By augmenting known physical laws with a trainable neural residual, our model captured the nuanced and localized deviations in battery evolution driven by heterogeneous load demand and noisy solar input.
The model exhibited stable and accurate predictions over both the training period and a three-times longer forecast horizon, affirming its potential for deployment in real-world energy systems. Our approach also showcased how modest variations in nodal consumption patterns could be learned without architectural complexity, highlighting the scalability of the framework to larger grids.
This work contributes to the growing body of research on hybrid modeling in scientific machine learning, particularly in the context of sustainable energy. By uniting interpretability from physical models with the expressive power of neural networks, UDEs offer a compelling toolkit for data-driven infrastructure modeling, energy forecasting, and control.
Future extensions may incorporate real-world data, multi-modal sensing (e.g., weather, market rates), and online control strategies, positioning this methodology as a viable candidate for next-generation smart grid management.

\bibliographystyle{IEEEtran}

\end{document}